\documentclass[journal,onecolumn]{IEEEtran}
\usepackage{authblk}
\usepackage{amsmath}
\usepackage{graphicx}
\usepackage{hyperref}
\usepackage{cite}
\usepackage{xurl}

\title{EVALUATION OF ARTIFICIAL INTELLIGENCE METHODS FOR LEAD TIME PREDICTION IN NON-CYCLED AREAS OF AUTOMOTIVE PRODUCTION}
\author[1,2]{Cornelius Hake}
\author[1,3]{Jonas Weigele}
\author[3]{Frederik Reichert}
\author[2]{Christian Friedrich}
\affil[1]{Dr. Ing. h.c. F. Porsche AG, Stuttgart, Germany, \authorcr \texttt{cornelius.hake1@porsche.de}}
\affil[2]{Hochschule Karlsruhe -- University of Applied Sciences, Karlsruhe, Germany}
\affil[3]{Hochschule Esslingen -- University of Applied Sciences, Esslingen, Germany}

\begin{document}
\maketitle

\begin{abstract}
The present study examines the effectiveness of applying Artificial Intelligence methods in an automotive production environment to predict unknown lead times in a non-cycle-controlled production area. Data structures are analyzed to identify contextual features and then preprocessed using one-hot encoding. Methods selection focuses on supervised machine learning techniques. In supervised learning methods, regression and classification methods are evaluated. Continuous regression based on target size distribution is not feasible. Classification methods analysis shows that Ensemble Learning and Support Vector Machines are the most suitable. Preliminary study results indicate that gradient boosting algorithms LightGBM, XGBoost, and CatBoost yield the best results. After further testing and extensive hyperparameter optimization, the final method choice is the LightGBM algorithm. Depending on feature availability and prediction interval granularity, relative prediction accuracies of up to 90\% can be achieved. Further tests highlight the importance of periodic retraining of AI models to accurately represent complex production processes using the database. The research demonstrates that AI methods can be effectively applied to highly variable production data, adding business value by providing an additional metric for various control tasks while outperforming current non AI-based systems.
\end{abstract}


\section{INTRODUCTION AND LITERATURE REVIEW}
Production planning and control (PPC) is an essential component for controlling and operating a production system. In contrast to vehicle assembly, the vehicles in the Test and Finish Centre (TFC) are no longer cycled due to the individual scopes of testing and work. As a result, the sequence of vehicles can be arbitrary, which leads to a need to optimize the sequence in PPC. The control of a TFC is an extremely complex task that depends on many influencing factors, especially in the production of vehicles with a high number of variants and a low number of units. For this reason, the exact lead time of a vehicle through the TFC cannot be determined with certainty at the present time \cite{KROPIK}.

In this context, the term lead time is defined as the interval between the sixth and eighth reporting points, which are assigned when a defined change of location of the vehicle occurs within the production chain. The sixth reporting point marks the conclusion of assembly and the transfer of the vehicle to the TFC, while the eighth reporting point signifies the completion of the vehicle and the transfer of ownership from the production to the sales department \cite{KROPIK}. The ability to predict the period between the sixth and eighth reporting points with a high degree of accuracy presents several potential advantages. From the perspective of the TFC, a further parameter is established, which serves to optimize the decision-making process for the scheduling and retrieval of individual vehicles. Furthermore, customers benefit from the enhanced precision of delivery dates. From the standpoint of sales, there is the potential for optimization in the planning of transportation. The explainable decision-making process of the AI model enables a root cause analysis, which provides information on the characteristics that are currently decisive in determining whether a vehicle has a longer or shorter lead time through the TFC. This offers the potential to optimize the TFC processes in a subsequent step.

The results of the literature research show that there are several works that deal with the prediction of lead times in production based on AI. It is important to note, that only the author's company own past research and that of Krzoska et al. specifically consider lead times in automotive production \cite{KRZOSKA}. Nevertheless, all studies pursue the goal on the precise prediction of lead times in industrial production using AI methods. 

Previous research by the author's company investigated two methods for predicting lead times in a TFC with AI: regression and classification using LightGBM (LGBM). The contextual features analyzed included four configuration variants (order type, color selection, steering wheel position, and special refinement) and all inspection codes recorded before the sixth reporting point. However, the limited selection of configuration variants reflects subjective feature relevance, potentially missing complex relationships in production. While LGBM can independently select relevant features, the previous approach did not fully exploit this potential. The sufficiency of the four configuration variants and the inspection codes as features is questionable, as shown by the model training results in Table 1. The LGBM regression achieved a prediction accuracy of 22\%, while the LGBM classification accuracy ranged between 65\% and 71\%, depending on the class distribution for the time intervals. No relevant optimization of the LGBM hyperparameters was performed. The results indicate that AI can predict vehicle lead times in a TFC, but the current results are not sufficient for series application. Further optimization and research on feature selection, method selection, and model-specific hyperparameters are needed.

\begin{table}[htbp]
\centering
\caption{Previous research results regarding class distribution and prediction accuracy}
\begin{tabular}{|c|c|c|c|}
\hline
\textbf{Algorithm} & \textbf{Method} & \textbf{Time Intervals [d]} & \textbf{Acc [\%]} \\ \hline
LGBM & Regression     & $<1$; $1{-}3$; $>3$            & 22 \\ \hline
LGBM & Classification & $<1$; $1{-}3$; $>3$            & 69 \\ \hline
LGBM & Classification & $<1$; $1{-}2$; $>2$            & 71 \\ \hline
LGBM & Classification & $<1$; $1{-}3$; $3{-}7$; $>7$   & 65 \\ \hline
\end{tabular}
\end{table}

The study from Krzoska et al. focuses on the prediction of rework times in automotive production. The aim is not to predict the total lead time of vehicles in a TFC, but rather the individual rework times in the TFC to rectify specific vehicle faults. To this end, process-related quality data from a MES is analyzed and the prediction of rework times is tested using vehicle-specific regression models. The regression tree methods Reduced Error Pruning Tree (REPtree) and M5 Model Trees (M5P) were analyzed. Results show that simply including MES data is insufficient, with prediction accuracy ranging from 48.3\% to 58.3\%. Manually removing outliers improves the accuracy from 64.5\% to 69.6\%. Introducing a severity rating factor can further improve accuracy to 84.9\%. REP-tree consistently outperforms M5P. However, the use of a weighting factor involves subjective human judgment and risks system manipulation. The study only considers individual rework times and shows relative errors up to 23.7\%. For cumulative errors, all processing times must be summed to predict total lead time. Further research is needed for AI methods to accurately predict total lead time in a TFC \cite{KRZOSKA}.

Further work in this field relates to other industries to predict lead times in production. Matzka \cite{MATZKA} and von der Hude \cite{HUD} discuss two different approaches to realize this. On the one hand, it is possible to use a trained regression model to form exact and continuous value predictions for the lead time. On the other hand, discrete predictions can be made in the form of time intervals using classification methods.

The studies by Krzoska et al. \cite{KRZOSKA}, Lingitz et al. \cite{LINGITZ}, Murphy et al. \cite{MURPHY}, Barzizza et al. \cite{BARZIZZA}, Aslan et al. \cite{ASLAN} and Gyulai et al. \cite{GYULAI} exclusively examine regression methods for the formation of lead time predictions. The studies by Lim et al. \cite{LIM} and Welsing et al. \cite{WELSING} are limited to research in the field of classification methods. The study by Bender \cite{BENDER} examines both regression and classification methods. Regardless of regression or classification, ensemble learning methods, such as random forest or gradient boosting, achieve the best results in six out of nine papers. Only the studies conducted by Krzoska et al. \cite{KRZOSKA} and Welsing et al. \cite{WELSING} have demonstrated that regression trees and cross-validated decision trees are the most effective in terms of prediction accuracy. The study by Lim et al. \cite{LIM} reaches the conclusion that Support Vector Machines (SVM) result in superior prediction outcomes in comparison to the other AI models. The work by Barzizza et al. \cite{BARZIZZA} demonstrates the potential of combining clustering with machine learning models, particularly Random Forest, to improve the accuracy of lead time predictions. Krätschmer \cite{KRATSCHMER} and Bentéjac et al. \cite{BENTEJAC} highlight the relevance of optimizing model-specific hyperparameters to improve the performance and generalization capability of AI models. However, only the studies by Krzoska et al. \cite{KRZOSKA} and Lingitz et al. \cite{LINGITZ} take this into account in their investigations. In general, the studies by Barzizza et al. \cite{BARZIZZA}, Aslan et al. \cite{ASLAN}, Murphy et al. \cite{MURPHY} and Gyulai et al. \cite{GYULAI} show that AI models are superior to traditional systems for predicting production lead times.

In this study, two major contributions are made to the existing literature on the subject:
\begin{enumerate}
    \item Method development for AI-based prediction of lead times for vehicles produced in low unit numbers and individual test scopes.
    \item Comparison of AI-based and current rule-based prediction of lead times.
\end{enumerate}

\section{METHODS AND MATERIALS}
\subsection{Experimental Setup}
Python is used as the programming language for the algorithms, which are computed on a workstation with an Intel Core i9-12900KF processor (16 cores, max. 5.20 GHz), 64 GB RAM, and an Nvidia RTX 3060 GPU with 12 GB GDDR6 graphics memory \cite{INTEL, NVIDIA}. To perform the analysis, it is necessary to collect, store and access the production data for the AI application, as illustrated in Figure 1.

\begin{figure}[htbp]
\centering
\includegraphics[width=0.6\textwidth]{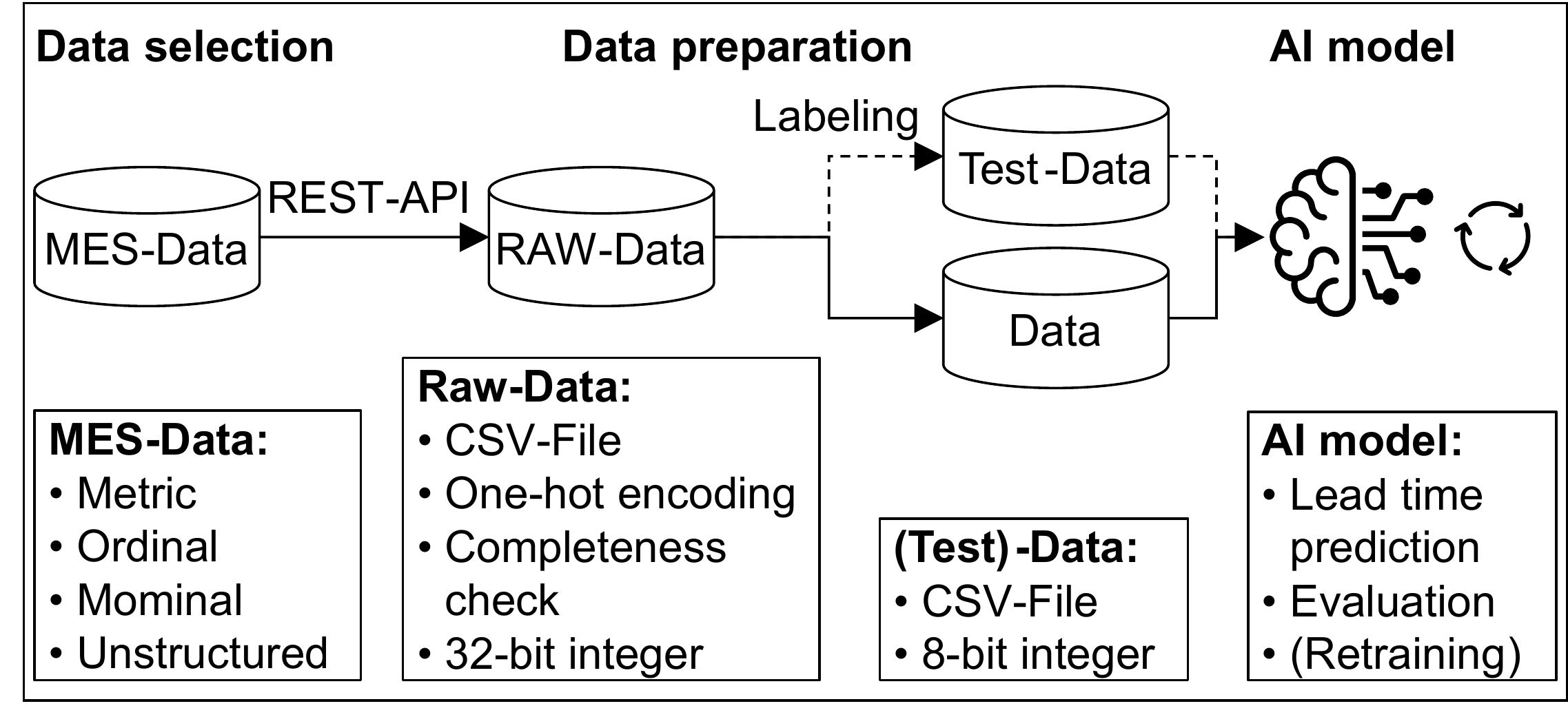}
\caption{Illustration of the data pipeline; dashed line: test, continuous line: series application}
\label{fig:data_pipeline}
\end{figure}

It is important to use a unique identification key with a vehicle reference to link data from various sources and assign it to a specific vehicle. The data can be available in metric, ordinal, nominal and unstructured form in each source, which is why standardized data preparation must be carried out for further processing.

\subsection{Obtaining and Preprocessing the Training Data}

\begin{figure}[htbp]
\centering
\includegraphics[width=0.6\textwidth]{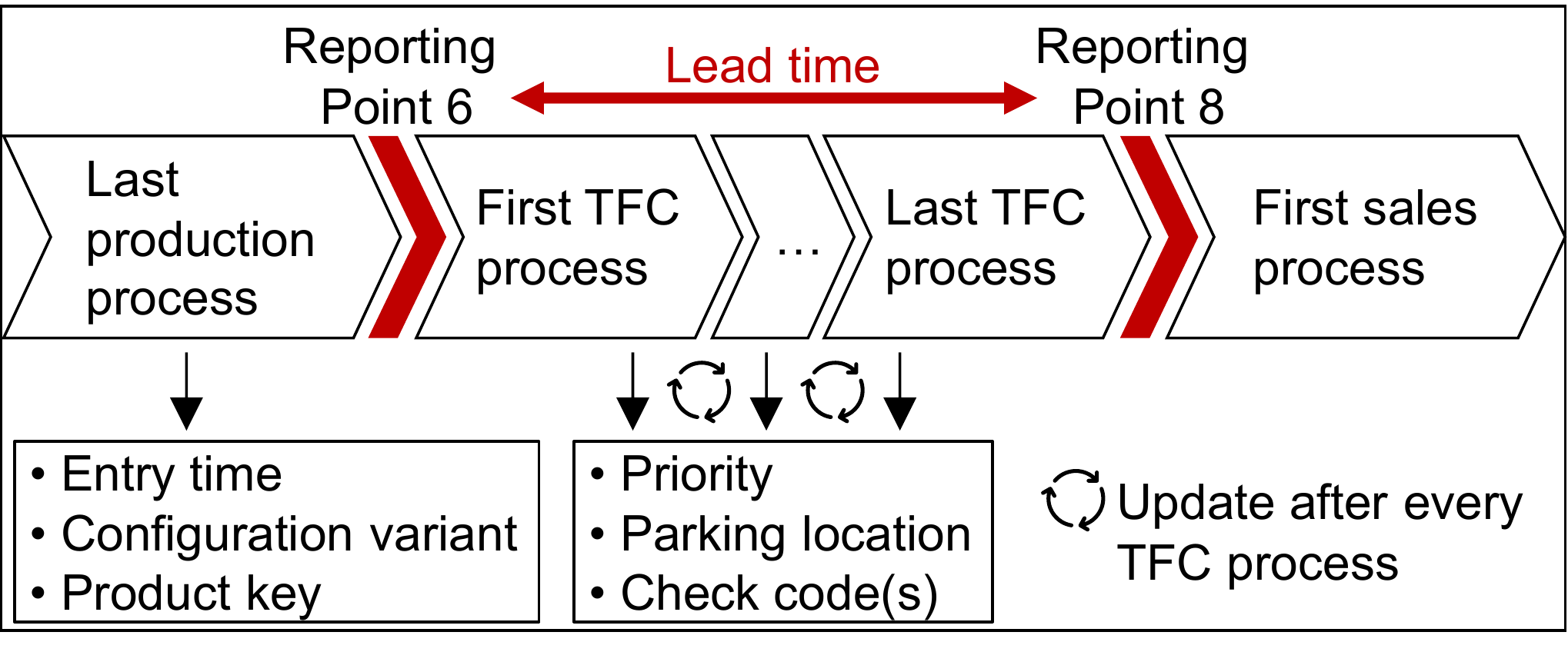}
\caption{TFC process with explanation of data availability and lead time calculation}
\label{fig:tfc_process}
\end{figure}

The first step is to select contextual data. A data model is searched for data influencing the vehicle lead time in the TFC. A distinction is made between characteristics known at the time of TFC entry and those recorded during the process. Known characteristics at entry include configuration variants, specific product key (vehicle derivative), and entry time (weekday and hour) as illustrated in Figure \ref{fig:tfc_process}. These define product complexity and shift time, considering shift changes and downtime. Additional characteristics captured during progression include priority, parking locations, and inspection type/location combination. Specific inspection codes for each vehicle influence the processing time in the TFC. Table \ref{tab:characteristics} provides an overview of relevant characteristics, data types, and processing types. Non-series vehicles are excluded from the analysis. The product key distinguishes between internal combustion engine vehicles (ICE) and battery electric vehicles (BEV) data.

\begin{table}[htbp]
\centering
\caption{Overview of data characteristics, their distribution per vehicle after TFC and preparation techniques}
\begin{tabular}{|c|c|c|c|c|}
\hline
\textbf{Characteristic} & \textbf{Data Type} & \textbf{Preparation Method} & \textbf{Example} & \textbf{n per Vehicle} \\ \hline
Priority & structured, ordinal & not necessary & 0 or 1 & 1 \\ \hline
Inspection code & structured, nominal & One-Hot-Encoding & 1234-9999F & 9422 \\ \hline
Parking location & unstructured, text & One-Hot-Encoding & Parking01-Level01 & 68 \\ \hline
Product key & structured, nominal & One-Hot-Encoding & A12BC & 1 \\ \hline
Entry time & structured, nominal & One-Hot-Encoding & Friday-19 & 1 \\ \hline
Configuration variant & structured, nominal & One-Hot-Encoding & AA1 & 2673 \\ \hline
\end{tabular}
\label{tab:characteristics}
\end{table}

Raw data includes both structured and unstructured forms. Since ML algorithms cannot directly process unstructured data, data preparation is essential. Features must remain unbiased by additional information. One-hot encoding is used to resolve this problem. One-hot encoding creates a unique, binary data set that allows all ML algorithms to process the same data set, making the computational results comparable. One-hot encoding increases the dimensionality of the data set. During dimension reduction, columns filled with null values are removed as they are irrelevant, thus reducing the dimensionality. An upstream check ensures the completeness of data points in the exported CSV files, and vehicles with incomplete characteristics are removed \cite{KAUPP}.

The data preparation described above is used to prepare several data sets for the training and testing process. The AI models to be investigated are trained with all data sets presented in Table \ref{tab:data_records}. The cases of feature availability are the known features when entering the TFC (data set = Limited) and all available features when passing through the TFC (data set = Unlimited). For the different production lines, a distinction is made between ICE and BEV vehicles. For development reasons, a common data set is also created for both production lines (Derivative = All).

\begin{table}[htbp]
\centering
\caption{Overview of data records by derivatives with the number of characteristics and data points}
\begin{tabular}{|c|c|c|c|c|}
\hline
\textbf{Data set} & \textbf{Derivative} & \textbf{Abbreviation} & \textbf{n Features} & \textbf{n Vehicles} \\ \hline
Limited & All & LA & 1,326 & 80,495 \\ \hline
Limited & ICE & LI & 1,326 & 45,706 \\ \hline
Limited & BEV & LB & 1,326 & 34,789 \\ \hline
Unlimited & All & UA & 12,166 & 80,495 \\ \hline
Unlimited & ICE & UI & 12,166 & 45,706 \\ \hline
Unlimited & BEV & UB & 12,166 & 34,789 \\ \hline
\end{tabular}
\label{tab:data_records}
\end{table}

\subsection{Selection Process and Methodology for AI Model Evaluation}

Two prediction approaches can be considered \cite{POKORNI}: exact continuous value prediction using regression models and discrete prediction using classification methods \cite{MATZKA, HUD}. Regression models with supervised learning require a linear data distribution, but since TFC lead times follow an exponential distribution, linear regression is inappropriate. Non-linear regression models are excluded due to their high resource requirements. Classification methods considered include decision trees, ensemble learning, naive Bayes, k-nearest neighbor, SVMs, and neural networks. Neural networks are resource intensive and lack explanatory power, while naive Bayes assumes feature independence, leading to poor generalization. K-nearest neighbor requires all data points, resulting in high resource requirements and longer classification times. Ensemble learning methods and SVMs are suitable due to good generalization and fast classification times. However, ensemble learning with decision trees loses explainability with more trees. Nevertheless, ensemble learning performs feature selection during training, which improves model performance and allows feature relevance analysis \cite{KAUPP, MATZKA, KNUTH}. A preliminary experiment is necessary to make the final selection decision. For this purpose, the specific methods of ensemble learning, and SVMs are first selected below \cite{KRECH}.

\begin{figure}[htbp]
\centering
\includegraphics[width=0.6\textwidth]{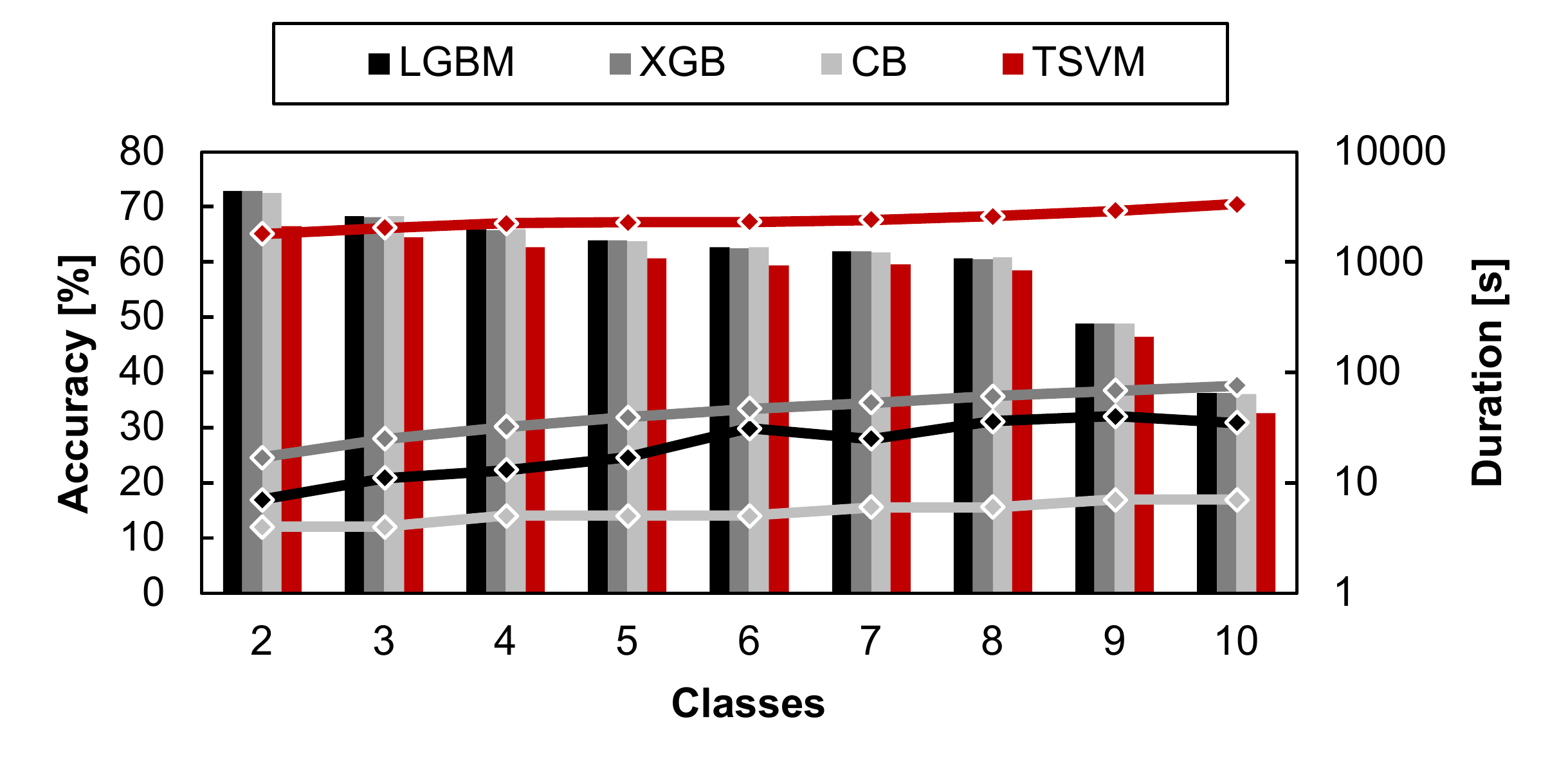}
\caption{Results of the prediction accuracy (bins) and training times (lines) in the preliminary experiment}
\label{fig:result_figure_pe}
\end{figure}

Ensemble learning methods such as random forest (bagging) and gradient boosting machines (boosting) are particularly suitable. Bagging reduces model variance by aggregating multiple independent models but can introduce bias and reduce prediction accuracy \cite{BOTSCH}. Boosting algorithms, such as LGBM, XGBoost (XGB), and CatBoost (CB), address this by weighting observations. Bentéjac et al. \cite{BENTEJAC} highlight these algorithms for their generalizability, training performance, and predictive precision. Therefore, the analysis focuses on these specific ML algorithms regarding their suitability for the prediction of lead times \cite{KE, CHEN, PROKHORENKOVA}. For configuring SVMs, choosing the kernel and strategy for multi-class classifications is crucial. The RBF kernel is ideal for complex data requiring non-linear separation. One-vs-All or One-vs-One methods are used, depending on the SVM library. ThunderSVM (TSVM), which supports OVO for multi-class classification, is suitable for GPU-based computations with the given setup. 
Figure \ref{fig:result_figure_pe} illustrates the results of the prediction accuracy (bins) and training times (lines) in the preliminary experiment.

In a preliminary experiment on the LA data set, LGBM, XGB, and CB achieved up to 73\% accuracy, while TSVM reached only 66\% as illustrated in Figure \ref{fig:result_figure_pe}. Gradient boosting machines trained faster and performed better, with TSVM taking 3348 s to train, 100 times slower than GBDT models. Due to suboptimal results, TSVM is excluded from further analysis.

\subsection{Labeling the Data}

In the supervised learning methodology, each data point requires an associated target variable or label that represents the expected output based on its attributes. For discrete labeling of the data points, groups are formed based on the actual lead times of the vehicles. Using domain-specific knowledge, the groups are defined based on time intervals by analyzing frequency distributions and the general relevance for the company to assign the data points \cite{KAUPP}. To predict the lead times between the sixth and eighth reporting points, both the operational benefit and sufficient data distribution within the classes must be considered. The exact time intervals of the classes are shown in Table \ref{tab:time_intervals}. Initially, nine test series are formed with an increasing number of classes. These start with two different classes and extend up to ten different classes. The classification for the lead time prediction starts with two classes: Vehicles that reach the eighth reporting point within 24 h from the sixth reporting point (55.7\%), and those that do so after 24 h (44.3\%). With increasing class division, the class with the most data points is subdivided more finely. With eight classes, except for the direct runners, no class contains more than 10\% of the data. The labels must be assigned to each data set as a target value for the training process.

\begin{table}[htbp]
\centering
\caption{Distribution of classes and the corresponding time intervals}
\begin{tabular}{|c|c|c|c|c|c|c|c|c|c|c|c|}
\hline
\textbf{Number of Classes} & \multicolumn{10}{c|}{\textbf{Intervals [d]}} \\ \hline
 & 1 & 2 & 3 & 4 & 5 & 6 & 7 & 8 & 9 & 10 \\ \hline
2 & [$\leq$1] & [$>$1] &  &  &  &  &  &  &  &  \\ \hline
3 & [$\leq$1] & ]1-3] & [$>$3] &  &  &  &  &  &  &  \\ \hline
4 & [$\leq$1] & ]1-3] & ]3-7] & [$>$7] &  &  &  &  &  &  \\ \hline
5 & [$\leq$1] & ]1-3] & ]3-7] & ]7-28] & [$>$28] &  &  &  &  &  \\ \hline
6 & [$\leq$1] & ]1-3] & ]3-7] & ]7-28] & ]28-84] & [$>$84] &  &  &  &  \\ \hline
7 & [$\leq$1] & ]1-2] & ]2-3] & ]3-7] & ]7-28] & ]28-84] & [$>$84] &  &  &  \\ \hline
8 & [$\leq$1] & ]1-2] & ]2-3] & ]3-7] & ]7-14] & ]14-28] & ]28-84] & [$>$84] &  &  \\ \hline
9 & [$\leq$0.5] & ]0.5-1] & ]1-2] & ]2-3] & ]3-7] & ]7-14] & ]14-28] & ]28-84] & [$>$84] &  \\ \hline
10 & [$\leq$0.25] & ]0.25-0.5] & ]0.5-1] & ]1-2] & ]2-3] & ]3-7] & ]7-14] & ]14-28] & ]28-84] & [$>$84] \\ \hline
\end{tabular}
\label{tab:time_intervals}
\end{table}

\subsection{Training Procedure for AI Methods}

This section describes the preparation, training, and validation of gradient boosting machines. First, a preliminary experiment with standard hyperparameters determines achievable target classes for which complex hyperparameter optimization is performed later. The data set is partitioned into 80\% training, 10\% validation, and 10\% test data, ensuring that each class is equally represented by stratified splitting and using the random state value 42 for consistency. After data preparation, model-specific default hyperparameters are defined, including multi-class classification via loss functions. Models are initialized with developer-defined default hyperparameters. The training process records training time, feature relevance, number of decision trees, and the AI model. To improve the performance, robustness, and generalization of GBDT models, it is recommended to optimize model-specific hyperparameters such as learning rate from 0.01 to 0.3, number of leaves from 32 to 1024, max depth from 2 to 200, leaf est. iterations from 1 to 10, and leaf reg. from 1 to 10 \cite{HUD}. This involves testing different hyperparameter combinations using grid search and k-fold cross-validation, and then evaluating their impact on training results. Due to the high resource requirements, it's performed on the LA and UA data sets, with the results applied to the ICE and BEV data sets. A fivefold cross-validation grid search is used, with 80\% of the data for training and 20\% for evaluation. Accuracy is chosen as the evaluation metric for the cross-validation results.

\section{RESULTS}
\subsection{Training Results}

Finally, LGBM, XGB, and CB are trained and tested for prediction accuracy. The test data labels are predicted by each model, and the actual labels are used to calculate the percentage of correct predictions, resulting in a relative prediction accuracy. Models that achieve at least 80\% prediction accuracy on the test data are further evaluated. The evaluation includes parameters such as prediction accuracy, training time, and number of decision trees generated.

The analysis shows that within the UA data set, the experimental design goals are only met for class numbers between two and four with prediction accuracies up to 90\%. For finer classifications from nine to ten classes, the prediction accuracy decreases significantly to 32\%. The LGBM model slightly outperforms the XGB and CB models in quality by two to four classes, using 460 instead of 999 (XGB) or 997 (CB) decision trees, highlighting its efficiency. In terms of training times, XGB takes 141 s, while CB is the fastest at 9 s, with LGBM also performing well at 16 s. To efficiently meet the test objectives, target classes must be limited to two to four, highlighting the predictability of wider intervals. Overall, the LGBM model leads in prediction accuracy and efficiency.

The analysis of the training results of the gradient boosting machines using the LA data set indicates an adjustment of the target value of the relative prediction accuracy to 70\% due to the reduced number of features. The results reveal that the target value is only achieved with a classification into two classes. Here, all three models examined show almost equivalent performance in terms of relative prediction accuracy up to 73\%. Here too, the LGBM model requires 308 instead of 838 (XGB) or 711 (CB) decision trees for comparable prediction performance. The evaluation of the training times is negligible since all models were trained in less than 20 s. In all cases, the F1 score reaches values above 0.92. From the synergy of reduced number of decision trees, solid prediction accuracy and short training time, the LGBM model emerges as the best in this comparative analysis.

The results show that the desired accuracy is only achieved for class numbers of two to four using the UA, UI, and UB data sets. For the data sets LA, LI, and LB, the prediction target is only achieved for two classes. Based on these findings, the resource-intensive hyperparameter optimization is only performed for those data sets for which the determined number of classes is available. Hyperparameter optimization of the LGBM, XGB, and CB models yields different optimal values. For LGBM and XGB, the default learning rates are generally optimal, while CB requires different combinations for different data sets. Tightening the tree evolution constraint helps prevent overfitting. K-fold cross-validation results show comparable performance for all models, with LGBM and XGB slightly outperforming CB. Training times were not compared because the optimization was performed only once.

The k-fold cross-validation results for the LA data set in Table \ref{tab:cross_validation} show that all models perform similarly with minor deviations. LGBM has a deviation of ± 1.2\%, XGB ± 1.4\%, and CB ± 1.3\%. LGBM and XGB achieve an average result of 71.3\% accuracy, slightly better than CB's 71.0\% accuracy. In conclusion, all three models show good generalization capabilities, with LGBM performing best overall.

\begin{table}[htbp]
\centering
\caption{Results in form of the accuracy in \% of the k-fold cross-validation for each GBDT model}
\begin{tabular}{|c|c|c|c|c|c|c|}
\hline
\textbf{Model} & \textbf{k = 1} & \textbf{k = 2} & \textbf{k = 3} & \textbf{k = 4} & \textbf{k = 5} & \textbf{Average} \\ \hline
LGBM & 71.7 & 72.5 & 70.0 & 70.5 & 72.1 & 71.3 \\ \hline
XGB & 71.8 & 72.7 & 69.9 & 70.2 & 72.2 & 71.3 \\ \hline
CB & 71.8 & 72.3 & 69.9 & 69.9 & 71.1 & 71.0 \\ \hline
\end{tabular}
\label{tab:cross_validation}
\end{table}

\subsection{Evaluation of the Prediction Quality}

\begin{figure}[htbp]
\centering
\includegraphics[width=0.6\textwidth]{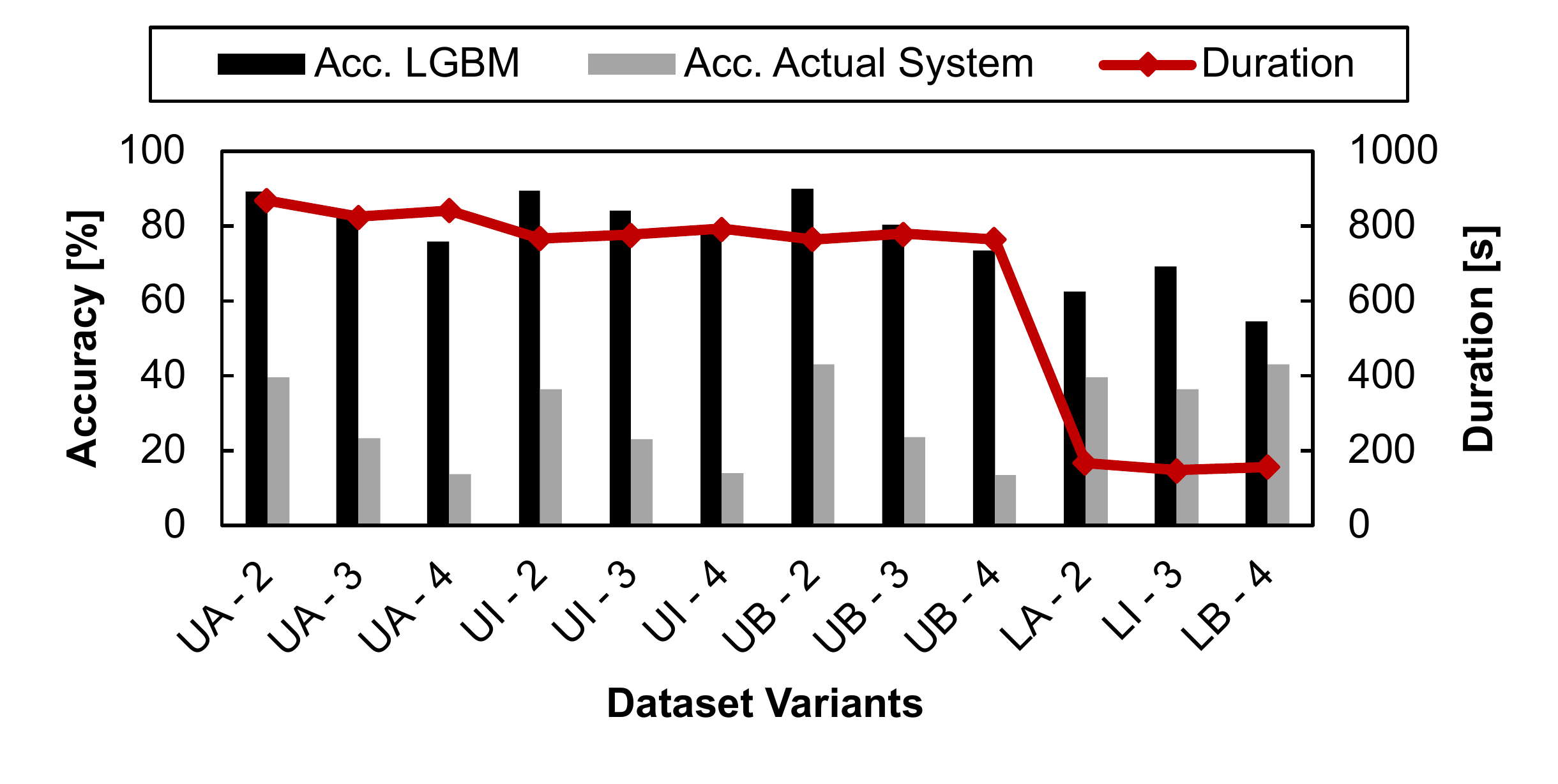}
\caption{Results from the comparison with the actual, rule-based system}
\label{fig:final_comparison}
\end{figure}

A comparison is made between the actual, rule-based and target, AI-based systems for 465 vehicles. The planned date of the actual system is created once before it is entered into the TFC and is not continuously updated as in the target system. As a result, LGBM achieves up to 65\% higher prediction accuracy than the actual system for data records without characteristic limits as it is illustrated in Figure \ref{fig:final_comparison}. The systems are best compared using the feature-limited data sets LA, LI, and LB, as in these cases the systems work with the same data basis. The results show that the LGBM model of the ICE production line achieves 32.8\% better prediction accuracy than the actual system with a data preparation and classification time of 146 s. In the case of the BEV production line, a prediction accuracy 11.5\% better than the actual system can be achieved with a data preparation and classification time of 156 s. The lower accuracy value is due to process-related changes implemented during this work. The test results of this comparison illustrate how powerful AI models can be compared to conventional systems. In all cases, the use of AI can achieve a significantly higher prediction accuracy of vehicle lead times in the TFC.

\section{CONCLUSION AND FUTURE WORK}

The research results show that AI-based prediction of lead times with GBDT models is possible for vehicles produced in low unit numbers and individual test scopes. The work also highlights the limitations of the system. Due to a process change in the BEV production line, the training data no longer accurately reflects the real production processes. As a result, the test results of the GBDT model are worse than the development results, which shows the high requirements for a good training database. The result illustrates the relevance of periodic retraining of the AI models when they are used in a real production environment. LGBM, as the GBDT model with the best test results, offers for the first time the ability to predict the lead time of a vehicle in the TFC with up to 90\% accuracy. This provides an additional decision-making parameter for production planning and control in the TFC. The test results show the improvement of the AI-based system compared to the actual, rule-based system. Using hyperparameter optimization, the average improvement in prediction accuracy was only 2\%.

A subsequent analysis of the relevance of the characteristics identified procedural inefficiencies. Two further research tasks arise from the research results. The first research task relates to the development of improved algorithms for explainable AI. Explainable decision-making enables a detailed analysis of the causes, creates transparency, and thus ensures that AI applications are widely accepted. The second research task is to develop an algorithm for the automated optimization of the time periods between the retraining of the AI models. The retraining ensures constant, high prediction quality. At the same time, training AI models is resource intensive and time consuming. Optimizing the number of retraining ensures that the prediction quality remains high while operating as economically as possible. This shows the relevance and motivation for further research in the field of AI applications in production.

\bibliographystyle{IEEEtran}
\bibliography{bibliography}

\end{document}